\documentclass[times,twocolumn,final,authoryear]{elsarticle}

\usepackage{ycviu}
\usepackage{hyperref}
\usepackage{bm}
\usepackage{framed,multirow}
\usepackage{amsmath}

\usepackage{amssymb}
\usepackage{bm}
\usepackage{url}
\usepackage{xcolor}
\definecolor{newcolor}{rgb}{.8,.349,.1}

\newlength\savewidth\newcommand\shline{\noalign{\global\savewidth\arrayrulewidth
  \global\arrayrulewidth 1pt}\hline\noalign{\global\arrayrulewidth\savewidth}}
\usepackage{tabularx} 
\usepackage{booktabs}

\usepackage{siunitx}  
\usepackage{cite}
\usepackage{tabularx}

\journal{Computer Vision and Image Understanding}

\begin{document}

\thispagestyle{empty}

\begin{frontmatter}

\title{\textsc{UAHOI}: Uncertainty-aware Robust Interaction Learning for HOI Detection}

\author[1]{Mu \snm{Chen}} 
\author[1]{Minghan \snm{Chen}}
\author[2]{Yi \snm{Yang}\corref{cor1}}
\cortext[cor1]{Corresponding author: Yi Yang}
\ead{yangyics@zju.edu.cn}

\address[1]{Australian Artificial Intelligence Institute, Faculty of Engineering and Information Technology, University of
Technology Sydney, NSW, Australia}
\address[2]{School of Computer Science, Zhejiang University, Zhejiang, China}

\received{1 May 2013}
\finalform{10 May 2013}
\accepted{13 May 2013}
\availableonline{15 May 2013}
\communicated{S. Sarkar}

\begin{abstract}
  This paper focuses on Human-Object Interaction (HOI) detection, addressing the challenge of identifying and understanding the interactions between humans and objects within a given image or video frame. Spearheaded by Detection Transformer (DETR), recent developments lead to significant improvements by replacing traditional region proposals by a set of learnable queries. However, despite the powerful representation capabilities provided by Transformers, existing Human-Object Interaction (HOI) detection methods still yield low confidence levels when dealing with complex interactions and are prone to overlooking interactive actions. To address these issues, we propose a novel approach \textsc{UAHOI}, Uncertainty-aware Robust Human-Object Interaction Learning that explicitly estimates prediction uncertainty during the training process to refine both detection and interaction predictions. Our model not only predicts the HOI triplets but also quantifies the uncertainty of these predictions. Specifically, we model this uncertainty through the variance of predictions and incorporate it into the optimization objective, allowing the model to adaptively adjust its confidence threshold based on prediction variance. This integration helps in mitigating the adverse effects of incorrect or ambiguous predictions that are common in traditional methods without any hand-designed components, serving as an automatic confidence threshold. Our method is flexible to existing HOI detection methods and demonstrates improved accuracy. We evaluate \textsc{UAHOI} on two standard benchmarks in the field: V-COCO and HICO-DET, which represent challenging scenarios for HOI detection. Through extensive experiments, we demonstrate that \textsc{UAHOI} achieves significant improvements over existing state-of-the-art methods, enhancing both the accuracy and robustness of HOI detection.
\end{abstract}

\begin{keyword}
\MSC 41A05\sep 41A10\sep 65D05\sep 65D17
\KWD Keyword1\sep Keyword2\sep Keyword3

\end{keyword}

\end{frontmatter}



\section{Introduction}

\begin{figure}[t]
  \vspace{-5pt}
  \begin{center}
      \includegraphics[width=1.\linewidth]{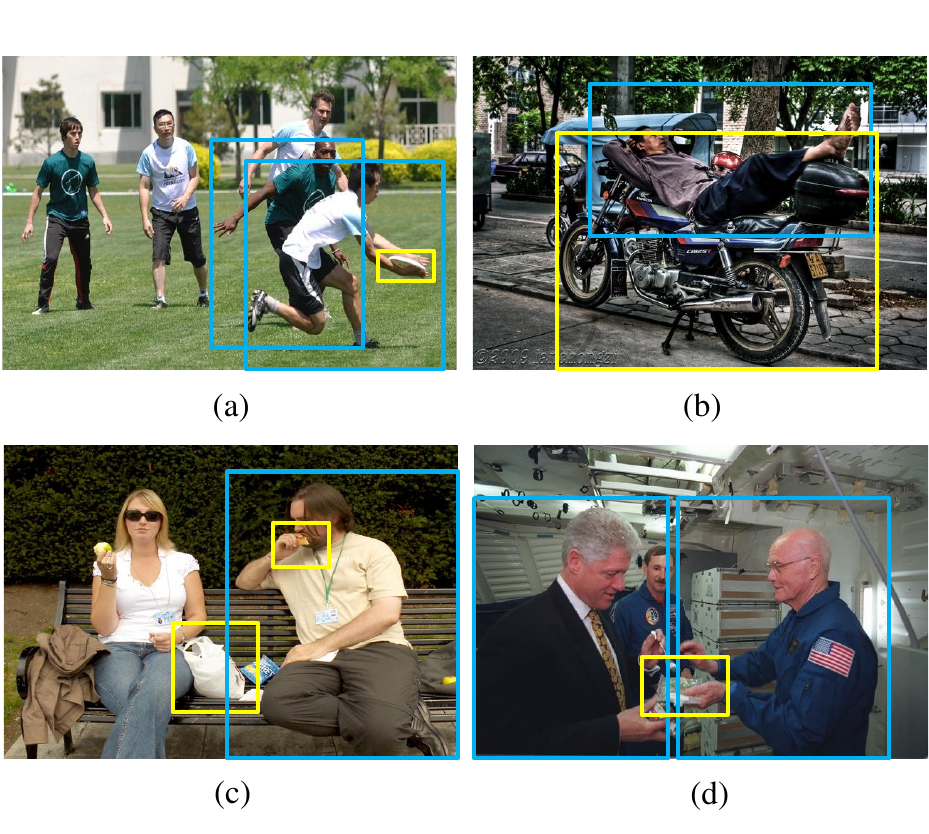}
      \end{center}
  \vspace{-20pt}
  \caption{Common challenges of current HOI Detection methods in complex scenes. The human/object bounding boxes are shown in blue/yellow.}
  \label{fig:1}
  \vspace{-15pt}
\end{figure}
Human-Object Interaction Detection (HOI Detection) aims to localize and recognize HOI triplets in the format of $<$human, verb, object$>$ from static images~\citep{ni2023human,leonardi2024exploiting}. This field stems from the detection of objects to include their relationships, prompting a deeper understanding on high-level semantic comprehension. HOI Detection has attracted considerable attention for its great potential in numerous high-level visual understanding tasks, including video question answering~\citep{wu2017visual}, video captioning~\citep{rao2024cmgnet,nian2017learning}, activity recognition~\citep{ozbulak2021investigating}, and syntia-to-reality translation~\citep{chen2023transferring}. 

Traditional methods~\citep{zhou2021cascaded,zhou2020cascaded,zhang2021spatially,gkioxari2018detecting} typically adopt either two-stage or one-stage pipeline, where the former detects instances first and then enumerates human-object paris to identify their interactions, and the latter attempts to do both simultaneously. However, these methods struggle with modeling the complex, long-range dependencies between humans and objects due to the localized nature of convolutional operations—a limitation that transformer-based methods address by capturing intricate interrelations across entire scenes. Recent advancements~\citep{li2023neural,chen2021reformulating,kim2021hotr,tamura2021qpic} have predominantly embraced the encoder-decoder framework pioneered by detection transformers (DETR)~\citep{carion2020end}, initializing learnable queries randomly, and subsequently decode the object queries into detailed triplets of human-object interactions. These methods offer enhanced accuracy by capturing global contexts and intricate interrelations, and simplify the architecture by eliminating the need for extra hand-designed components.

Despite the notable advances, several challenges still persist. As shown in Fig~\ref{fig:1}, \textbf{firstly}, among the wide range of predicates, some interactions may not be directly manifested through visual signals, often involving subtle movements and non-physical connections. For example, the interaction looking in a certain direction in (c) may easily be overlooked. And a man lying on a mortobike could be easily recognized as a man riding a mortobike (b). \textbf{Secondly}, in a multi-human scene, a person who may be imagining or planning to interact with an object without yet taking any action can easily be mistaken for having already performed the action, such as the unselected background individuals in images (a) and (d). \textbf{Thirdly}, the same type of interaction can appear very differently in different contexts. For instance, the action of ``grabbing" can vary significantly when interacting with different objects, depending on the object's size, shape, weight, and other characteristics. In such complex scenarios, current models typically assign lower confidence to interactions, which affect the performance. 

To address these issues, conventional two-stage methods involves assisstance from extra communication signal~\citep{shen2018scaling,li2023logicseg,liang2023logic,fan2019understanding,wang2019learning}, and language~\citep{xu2019learning,liu2020amplifying}, but the models still tend to focus on inaccurate regions. More recent approaches independently process instance detection and interaction classification using two separate decoders, which operate either in parallel~\citep{zhou2022human} or cascaded~\citep{zhang2021mining,chen2021reformulating} mode.
For example, \citet{zhang2021mining} applied two cascade decoders, one for generating human-object pairs and another for dedicated interaction classification of each pair, which helps the model to determine which regions inside a scene to concentrate on. Taking a step further, Unary~\citep{zhang2022efficient} separately encodes human and object instances, enhancing the output features with additional transformer layers for more accurate HOI classification. ~\citet{zhou2022human} disentangle both encoder and decoder to enhance the learning process for two distinct subtasks: identifying human-object instances and accurately classifying interactions, which necessitates learning representations attentive to varied regions. While the disentanglement of subtasks allows individual modules to concentrate on their specific tasks, thereby boosting overall performance, these methods often require additional heuristic thresholding when deciding which interactions to retain. For simple interactions, it is relatively easy to obtain good interaction predictions with high confidence scores in their predicted categories. However, for more complex interactions, confidence scores may be lower. In such cases, finding a suitable threshold manually to disregard low-confidence predictions could lead to overlooking correct interactions, representing a persistent challenge in HOI detection. Specifically, determining the optimal threshold value is challenging across different categories, and estimiting a value in advance is more complex. For overt interactions like ``riding," where a person is physically mounted on an object such as a bicycle or horse, confidence scores are typically high. In contrast, subtle interactions like ``reading", characterized by the presence of a book and the direction of a person's gaze, often yield lower overall confidence scores. A high threshold in such cases might cause the model to ignore these interactions. 

We propose a novel method \textsc{UAHOI}, Uncertainty-aware Robust Human-Object Interaction Learning that utilizes uncertainty estimation to dynamically adjust the threshold for interaction predictions in the HOI detection task. This approach integrates uncertainty modeling to refine the decision-making process, enabling the model to adjust its confidence thresholds based on the predicted uncertainty associated with each interaction. Specifically, we utilize the variance in predictions as a measure of uncertainty for both human/object bounding boxes and interaction, which reflects the model's confidence in their outputs. The variance is directly incorporated into our optimization target, enhancing the accuracy of bounding box predictions and ensuring that significant interactions are not overlooked due to artificially low confidence thresholds. Such adaptive handling of complex interactions increases the robustness of HOI detection models. \textsc{UAHOI} handles complex interactions in an adaptive manner, enhancing the accuracy of the bounding box predictions and preventing important interactions from being overlooked due to artificially low confidence thresholds, thereby increasing the robustness of HOI Detection model.
We conducted a comprehensive evaluation on two standard human-object interaction datasets HICO-DET~\citep{chao2018learning} and V-COCO~\citep{gupta2015visual}, and our experiments demonstrate a significant improvement over the existing state-of-the-art methods. Specifically, \textsc{UAHOI} achieved 34.19 mAP on HICO-DET and 62.6 mAP on V-COCO.

\section{Related Works}
\subsection{Traditional HOI Detection}
Human-Object Interaction (HOI) detection provides numerous high-level intricate relationships between humans and objects, gradually serving as the foundation of many computer vision applications~\citep{wang2024visual,chen2023pipa,chen2024pipa++,chen2024general}. Traditional Human-Object Interaction (HOI) detection methods can typically be divided into two categories: two-stage methods~\citep{qi2018learning,gupta2019no,wan2019pose,xu2019learning,zhou2019relation,li2019transferable,ulutan2020vsgnet,wang2020contextual,gao2020drg,hou2020visual,li2020hoi,kim2020detecting,li2020detailed}  and one-stage methods~\citep{kim2020uniondet,liao2020ppdm,wang2020learning,fang2021dirv}. Two-stage HOI detection methods rely on an off-the-shelf object detector to extract bounding boxes and class labels for humans and objects in the first stage. In the second stage, they model interactions for each human-object pair via a multi-stream network. For example, \citet{qi2018learning} propose leveraging the Graph Parsing Neural Network (GPNN) to incorporate structural knowledge into HOI detection. Similarly, \citet{gupta2019no} introduces a streamlined factorized model that utilizes insights from pre-trained object detectors. Subsequent research often involves integrating additional contextual and relational information to enhance performance further. However, two-stage methods find it challenging to identify human-object pairs among a large number of permutations and heavily rely on the detection results, suffering from low efficiency and effectiveness.\citep{liu2020amplifying,zhou2020cascaded,zhang2021spatially,gkioxari2018detecting,zhou2021cascaded}. By introducing anchor points to associate humans and objects, single-stage methods detect pairs likely to interact and their interactions simultaneously. This approach, which handles instance detection and interaction point prediction branches in parallel, has made impressive progress in HOI detection. For instance, \citet{kim2020uniondet} utilizes a union-level detector to directly capture the region of interaction, enhancing focus on interaction-specific areas. Meanwhile, \citet{liao2020ppdm} employs point detection branches that concurrently predict points for both the human/object and their interactions. This method not only implicitly provides context but also offers regularization for the detection of humans and objects, improving overall accuracy and context relevance. However, these one-stage methods still fall short in striking an appropriate balance in multi-task learning and modeling long-range contextual information.

\subsection{End-to-End HOI Detection}
Inspired by DETR~\citep{carion2020end}, recent work~\citep{chen2021reformulating,kim2021hotr,tamura2021qpic,zou2021end,zhang2021mining,kim2022mstr,zhang2022efficient,zhou2022human,liao2022gen,liu2022interactiveness,yuanrlip,zhong2022towards} modify HOI as a set-prediction problem by generating a set of HOI triplets. Early approaches~\citep{tamura2021qpic,zou2021end} simply migrated the Transformer decoder to HOI tasks, using a single decoder to couple human-object detection and interaction classification, achieving end-to-end training. \citet{tamura2021qpic} replaces manually defined location-of-interest with a transformer-based feature extractor, enhancing feature representation capabilities. \citet{zou2021end} addresses HOI detection using an end-to-end approach, which eliminates the reliance on hand-designed components, streamlining the detection process. However, due to the significant differences between the two tasks, learning a unified instance-interaction representation proved challenging. Therefore, subsequent works~\citep{kim2021hotr,zhang2021mining,chen2021reformulating} gradually shifted towards using separate decoders for instance prediction and interaction prediction, allowing the model to fully focus on the differences between instance and interaction prediction areas. To further enhance performance, other methods have introduced language~\citep{yuan2022detecting,cao2023detecting} and logic-reasoning~\citep{li2023neural} to explore the relationships between humans, objects, and their interactions more deeply. Moreover, scene graph generation (SGG) is another high-level semantic understanding task closely related to HOI detection, which is advanecd with the help of techniques such as noisy label correction~\citep{li2024nicest}, LLMs~\citep{li2023zero} and compositional augmentation~\citep{li2023compositional}. Both HOI Detection and SGG aim to capture and understand complex relationships between objects within a scene, but scene graph generation focuses on identifying objects and their pairwise relationships to create a structured graph representation, while HOI detection specifically targets interactions between humans and objects. Our method is built on a disentangled network framework that separately handles human-object detection and interaction classification tasks. 

\subsection{Uncertainty Estimation}
Our work is closely related to Uncertainty Estimation techniques~\citep{frankle2020linear}. The study of uncertainty estimation within deep learning frameworks is garnering considerable attention. Existing works have explored the uncertainty estimation from different aspects, such as Bayesian approaches~\citep{sensoy2018evidential,amini2020deep,gal2016dropout,gal2017concrete} and deep ensemble techniques~\citep{lakshminarayanan2017simple,vyas2018out}. Bayesian approaches adopt baysian theory to predict the uncertainty of the neural networks. Among them, \citet{gal2016dropout} and \citet{gal2017concrete} leavrage dropout to obtain probability distribution on network parameters, often come with a high computational cost and complexity. Ensemble methods aggregate predictions from multiple networks to estimate uncertainty, providing a balance between computational efficiency and reliability. Uncertainty estimation technologies make great progress in computer vision tasks such as semantic segmentation~\citep{zheng2021rectifying,lu2023uncertainty,fleuret2021uncertainty} and object detection~\citep{liu2016ssd,miller2018dropout,miller2019benchmarking}, but are less explored for HOI detection. In this work, we integrate uncertainty estimation into HOI detection by modeling bounding box coordinates (left, right, top, bottom) as Gaussian-distributed random variables and using dropout not only for regularization but also to estimate variance in interaction predictions. This allows us to embed uncertainty directly into the learning process, enhancing both the robustness and adaptability of our detection systems. Furthermore, we introduce an Uncertainty-Aware Loss Function and implement Regularization of Interaction Uncertainty as an automatic thresholding mechanism, demonstrating an effective approach to improving HOI detection through advanced uncertainty modeling techniques.

\section{Mehtods}

\begin{figure*}[t]
  \vspace{-5pt}
  \begin{center}
      \includegraphics[width=0.98\linewidth, height=210pt]{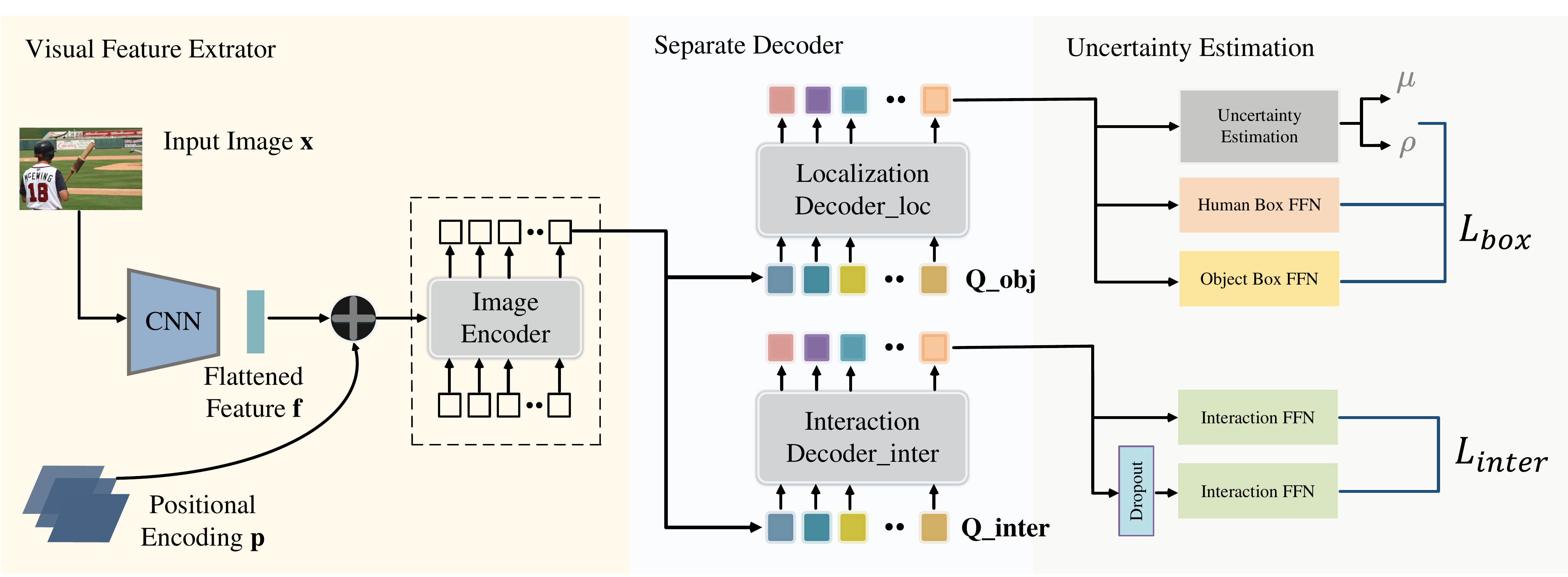}
      \end{center}
  \vspace{-10pt}
  \caption{Overall framework of our \textsc{UAHOI}. \textsc{UAHOI} consists of three components: Visual Feature Extrator, Parallel Decoder and Uncertainty Estimation module. Visual features are firstly extracted by CNN and shared Transformer Encoder. Then, the Localization Decoder and Interaction Decoder run n parallelto extract human/object bounding boxes and interaction class. Lastly, the proposed Uncertainty-aware Instance Localization and Interaction Refinement modules are used to perform uncertainty regularization.}
  \label{fig:2}
  \vspace{-10pt}
\end{figure*}

\subsection{Preliminary: Vanilla Transformer-based HOI Detection}
Since DETR~\citep{carion2020end}, object detection has been investigated as a set prediction problem. DETR employs a transformer encoder-decoder architecture to transform $N$ positional encodings into $N$ predictions, encompassing both object class and bounding box coordinates.
Similar to object detection, recent advancements\citep{chen2021reformulating,kim2021hotr,tamura2021qpic,zou2021end,zhang2021mining,kim2022mstr,zhang2022efficient,zhou2022human,liao2022gen,liu2022interactiveness,yuanrlip,zhong2022towards} have adeptly harnessed the Encoder-Decoder framework, integrating the Transformer architecture to better capture complex dependencies between humans and objects. This integration significantly enhances model precision and deepens understanding of interactions within a scene \citep{kim2021hotr,zou2021end}.
In our approach, as is shown in Fig~\ref{fig:2}, we implement an encoder-decoder structure with a shared encoder alongside two parallel decoders: one for instance localization and another for interaction recognition. This design helps to eliminate the issue of redundant predictions.~\citep{kim2021hotr}. In detail, the feature $\bm{f} \in \mathbb{R}^{D \times H \times W}$ is extracted from the input image $\bm{x}$ via a CNN backbone, where $H$ and $W$ are the size of the input image, and $D$ is the number of channel. Combined with positional embedding $p$, the feature $\bm{f}$ containing semantic concepts is flattened to construct a sequence of length $ H \times W$ and then fed into the image encoder. We adopt ResNet as our backbone. Each image encoder layer consists of a multi-head self-attention (MHSA) module and a feed-forward network, which refine the feature representation sequentially. After processing by image encoder, the resulting encoded features, denoted as $\bm{f}_{en} \in \mathbb{R}^{D^{\prime} \times H^{\prime} \times W^{\prime}}$, are split and fed to the instance localization decoder and interaction recognition decoder. The instance localization decoder identifies and localizes objects within the scene, and the interaction recognition decoder analyzes the interaction between detected objects and humans, aiming to understand their mutual interactions. Object queries $\bm{Q}_{obj}=\left\{\boldsymbol{q}_i \mid \boldsymbol{q}_i \in \mathbb{R}^d\right\}_{i=1}^H$ and Interaction queries $\bm{Q}_{inter}=\left\{\boldsymbol{q}_i \mid \boldsymbol{q}_i \in \mathbb{R}^d\right\}_{i=1}^H$ are initialized randomly, and then learnt via cross-attention layers. $H$ is the number of queries and $d$ is the query dimension.  The decoding process for localization and interaction recognition could be formulated by:
\begin{equation}
  \boldsymbol{L}=\operatorname{Decoder}_{l o c}\left(\boldsymbol{f}_{e n}, \boldsymbol{Q}_{o b j}\right) \in \mathbb{R}^{H \times 4},
\end{equation}

\begin{equation}
  \boldsymbol{I}=\operatorname{Decoder}_{\text {inter }}\left(\boldsymbol{f}_{\text {en }}, \boldsymbol{Q}_{\text {inter }}\right) \in \mathbb{R}^{H \times C}.
  \end{equation}
where $\boldsymbol{Q}_{obj}$ and $\boldsymbol{Q}_{\text {inter }}$ are sets of initialized queries for object detection and interaction recognition, respectively, refined through cross-attention mechanism.

With the learned HOI queries, the HOI pair could be decoded by several MLP branches. Specifically, we adopt three MLP branches designed to output the confidence levels for the human, object, and their interaction, respectively, each employing a softmax function to ensure probabilistic outputs. The addition of these branches allows for a more granular understanding of the HOI dynamics by providing individual confidence scores that reflect the certainty of each element's involvement in the interaction. The output embedding is then decoded into specific HOI instance via several multiple Multi-Layer Perceptron (MLP) layers. In detail, three separate MLP branches are designed to predict the confidence levels for the human, object, and interaction, respectively. Each branch employs a softmax function to generate probabilistic outputs. The human and object branches are denoted by orange color in Figure~\ref{fig:2}. For human branch, two values with confidence are outputed to indicate the likelihoods of foreground and background presence. Regarding object and interaction branches, the output scores including all categories of objects or actions and another one category for the background. \textsc{UAHOI} adopts two Fully Feed-Forward Networks (FFN) layers to predict the bounding boxes of the human and the object. The bounding box consists of four values to represent each coordinate for precise localization within the visual scene.

\subsection{Uncertainty-aware Instance Localization}
Firstly, when estimating localization uncertainty, we consider the bounding boxes for both humans, denoted by $C_{human} \in (l_h, r_h, t_h, b_h)$ and objects, represented as $C_{object} \in (l_o, r_o, t_o, b_o)$. This representation allows us to explore inherent uncertainty in the prediction of bounding box coordinates, which is particularly useful in complex scenes where occlusion or interaction may obscure part of the subjects. We employ a dedicated network to compute the standard deviations of these distributions, providing a measurable and quantifiable uncertainty which enhances the precision in object boundary detection This methodology not only improves accuracy but also increases the reliability of localizations by effectively capturing and quantifying the inherent uncertainties associated with positional offsets.
To accurately delineate the object's boundary, it is essential to account for the four directional offsets of the human/object bounding box. Adopting the framework outlined by~\citep{lee2022localization}, we implement an uncertainty estimation network tailored to assess the localization uncertainty derived from these regressed box offsets $(l, r, t, b)$, defined as Gaussian distributions:
\begin{align}
  l &\sim \mathcal{N}(\mu_l, \sigma_l^2), \\
  r &\sim \mathcal{N}(\mu_r, \sigma_r^2), \\
  t &\sim \mathcal{N}(\mu_t, \sigma_t^2), \\
  b &\sim \mathcal{N}(\mu_b, \sigma_b^2).
\end{align}
Here, $\mu$ and $\sigma^2$ signify the mean and variance of the offsets, respectively. To determine these parameters accurately, we deploy a neural network featuring a dual-head architecture. One head predicts the mean values while the other calculates the logarithm of the variance as \textit{uncertainty}, naming $Var_{box}$, ensuring that the variance $\sigma_{\text{box}}$ is always positive:

\begin{align}
  \mu_{\text{box}} &= \text{MLP}_{\mu}(\mathbf{f}_{\text{en}}), \\
  \sigma_{\text{box}}^2 &= \log(1 + \exp(\text{MLP}_{\sigma}(\mathbf{f}_{\text{en}}))).
\end{align}

Further, \citet{lee2022localization} introduce a Negative Power Log-Likelihood Loss, which is reformulated to achieve an uncertainty loss. This uncertainty loss compels the network to output a higher uncertainty value when the coordinate predictions from the regression branch are off-target:

\begin{equation}
  L_{box}=-\sum_{c \in\{l, r, t, b\}} IoU \cdot \log P_{\Theta}\left(C \mid \mu_c, \sigma_c^2\right).
\end{equation}
This equation emphasizes the integration of the Intersection over Union (IoU) as a scaling factor, where IoU measures the overlap between the predicted and actual bounding boxes, enhancing the training focus on precision. The probability density function $P_{\Theta}$, parameterized by network parameters $\Theta$, plays a crucial role in adjusting the model's certainty regarding the predicted localizations, thus pushing the boundaries of accuracy in object detection in highly dynamic and unpredictable scenes.

\subsection{Uncertainty-aware Interaction Refinement}
Secondly, to refine the interaction, we model the uncertainty of the interaction classification via the prediction variance. Typically, models tend to make less accurate predictions in complex interaction areas. By modeling uncertainty, we are able to quantitatively calculate this uncertainty. Specifically, if there is a significant difference between predictions made with and without dropout, the variance will be high. This reflects the model's uncertainty in predicting interactions. Following previous works~\citep{srivastava2014dropout}, we add structured noise to the interaction feature representation via dropout. We denote $\hat{\boldsymbol{y}}^i=f\left(\boldsymbol{x} ; \hat{\boldsymbol{\theta}}_i\right)$ and $\boldsymbol{y}=f(\boldsymbol{x} ; \boldsymbol{\theta})$ as the interaction representation with/without dropout. Following recent works in the field of uncertainty estimation~\citep{zheng2021rectifying}, we predict the interaction variance \textit{uncertainty} $Var_{inter}$ as the KL divergence between the two representation:

\begin{equation}
  D_{k l}=\mathbb{E}\left[\boldsymbol{y} \log \left(\frac{\boldsymbol{y}}{\hat{\boldsymbol{y}}^i}\right)\right].
  \end{equation}

Following ~\citep{kendall2017uncertainties,zheng2021rectifying}, we regularize the interaction variance by minimizing the prediction bias, thus enabling the learning from inaccurate interaction. The objective could be formulated as:

\begin{equation}
  L_{inter}=\mathbb{E}\left[\exp \left\{-D_{k l}\right\} L_{ce}+D_{k l}\right].
\end{equation}

The final loss to be minimized consists of four parts: 
\begin{equation}
  L_{\text {total }}=L_{loc}^h+L_{loc}^o+\lambda_o L_{box}+\lambda_a L_{inter}.
\end{equation}

Here, $\lambda_1$ and $\lambda_2$ are the weights of two uncertainty losses, $L_{loc}^h$ and $L_{loc}^o$ are computed by box regression loss. 
In this situation, during the optimization process, the variance of both bounding boxes and the interaction will be minimized.

\subsection{Implementation Details}
\noindent\textbf{Network Architecture.} To ensure a fair comparison with existing works~\citep{kim2021hotr,zou2021end,chen2021reformulating}, we adopt ResNet-50 as our backbone, followed by a six -layer transformer encoder as our visual feature extrator. Both the Localization and Interaction Decoder consist of four Transformer decoder layers. 

\noindent\textbf{Training.} During training, all of the transformer layer weights are initialized with Xavier init~\citep{glorot2010understanding}. \textsc{UAHOI} is optimized by AdamW~\citep{loshchilov2017decoupled} and we set the initial learning rate of both encoder and decoder to $10^{-4}$ and weight decay to $10^{-4}$. The weight coeffieicnts $\lambda_o$ and $\lambda_a$ are set to 1 and 1. To fairly compare with existing methods, the Backbone, Image Encoder and both Localization and Interaction Decoder are pretrained in MS-COCO and frozen during training. All the augmentation are the same as those in DETR~\citep{carion2020end}. All experiments are conducted on 8 A40 GPUs with a batch size of 16.

\section{Experimental Results}

\begin{table*}[t]\small
  \centering
  \setlength{\belowcaptionskip}{4pt}
   \caption{Comparison of detection performance on the HICO-DET~\citep{chao2018learning} and V-COCO~\citep{gupta2015visual} test sets, using ResNet50 backbone. The best performance is emphasized in bold.}
   \begin{tabularx}{\linewidth}{@{\extracolsep{\fill}} l l cccccccc}
      \toprule
    & & \multicolumn{6}{c}{\textbf{HICO-DET}} & \multicolumn{2}{c}{\textbf{V-COCO}} \\ [4pt]
    & & \multicolumn{3}{c}{Default Setup} & \multicolumn{3}{c}{Known Objects Setup} & & \\ 
    \cline{3-5}\cline{6-8}\cline{9-10} \\ [-8pt]
      \textbf{Method} & \textbf{Backbone} & Full & Rare & Non-rare & Full & Rare & Non-rare & AP$_{role}^{S1}$ & AP$_{role}^{S2}$ \\
      \midrule
     HO-RCNN~\citep{chao2018learning} & CaffeNet & 7.81 & 5.37 & 8.54 & 10.41 & 8.94 & 10.85 & - & - \\
     InteractNet~\citep{gkioxari2018detecting} & ResNet-50-FPN & 9.94 & 7.16 & 10.77 & - & - & - & 40.0 & - \\
     GPNN~\citep{qi2018learning} & ResNet-101 & 13.11 & 9.34 & 14.23 & - & - & - & 44.0 & - \\
     iCAN~\citep{gao2018ican} & ResNet-50 & 14.84 & 10.45 & 16.15 & 16.26 & 11.33 & 17.73 & 45.3 & 52.4 \\
     TIN~\citep{li2019transferable} & ResNet-50 & 17.03 & 13.42 & 18.11 & 19.17 & 15.51 & 20.26 & 47.8 & 54.2 \\
     Gupta et al~\citep{gupta2019no} & ResNet-152 & 17.18 & 12.17 & 18.68 & - & - & - & - & - \\
     VSGNet~\citep{ulutan2020vsgnet} & ResNet-152 & 19.80 & 16.05 & 20.91 & - & - & - & 51.8 & 57.0 \\
     DJ-RN~\citep{li2020detailed} & ResNet-50 & 21.34 & 18.53 & 22.18 & 23.69 & 20.64 & 24.60 & - & - \\
     PPDM~\citep{liao2020ppdm} & Hourglass-104 & 21.94 & 13.97 & 24.32 & 24.81 & 17.09 & 27.12 & - & - \\
     VCL~\citep{hou2020visual} & ResNet-50 & 23.63 & 17.21 & 25.55 & 25.98 & 19.12 & 28.03 & 48.3 & - \\
     ATL~\citep{hou2021affordance} & ResNet-50 & 23.81 & 17.43 & 27.42 & 27.38 & 22.09 & 28.96 & - & - \\
     DRG~\citep{gao2020drg} & ResNet-50-FPN & 24.53 & 19.47 & 26.04 & 27.98 & 23.11 & 29.43 & 51.0 & - \\
     IDN~\citep{li2020hoi} & ResNet-50 & 24.58 & 20.33 & 25.86 & 27.89 & 23.64 & 29.16 & 53.3 & 60.3 \\
     HOTR~\citep{kim2021hotr} & ResNet-50 & 25.10 & 17.34 & 27.42 & - & - & - & 55.2 & 64.4 \\
     FCL~\citep{hou2021detecting} & ResNet-50 & 25.27 & 20.57 & 26.67 & 27.71 & 22.34 & 28.93 & 52.4 & - \\
     HOI-Trans~\citep{zou2021end} & ResNet-101 & 26.61 & 19.15 & 28.84 & 29.13 & 20.98 & 31.57 & 52.9 & - \\
     AS-Net~\citep{chen2021reformulating} & ResNet-50 & 28.87 & 24.25 & 30.25 & 31.74 & 27.07 & 33.14 & 53.9 & - \\
     SCG~\citep{zhang2021spatially} & ResNet-50-FPN & 29.26 & 24.61 & 30.65 & 32.87 & 27.89 & 34.35 & 54.2 & 60.9 \\
     QPIC~\citep{tamura2021qpic} & ResNet-101 & 29.90 & 23.92 & 31.69 & 32.38 & 26.06 & 34.27 & 58.8 & 61.0 \\
     MSTR~\citep{kim2022mstr} & ResNet-50 & 31.17 & 25.31 & 32.92 & 34.02 & 28.82 & 35.57 & 62.0 & 65.2 \\
     CDN~\citep{zhang2021mining} & ResNet-101 & 32.07 & 27.19 & 33.53 & 34.79 & 29.48 & 36.38 & \textbf{63.9} & 65.9 \\
     UPT~\citep{zhang2022efficient} & ResNet-101-DC5 & 32.62 & 28.62 & 33.81 & 36.08 & 31.41 & 37.47 & 61.3 & \textbf{67.1} \\
     RLIP~\citep{yuanrlip} & ResNet-50 & 32.84 & 26.85 & 34.63 & - & - & - & 61.9 & 64.2 \\
     GEN-VLKT~\citep{liao2022gen} & ResNet-50 & 33.75 & 29.25 & 35.10 & 36.78 & 32.75 & 37.99 & 62.4 & 64.5 \\
    \midrule
    \textsc{UAHOI} & ResNet-50 & \textbf{34.19} & \textbf{31.54} & \textbf{35.27} & \textbf{37.44} & \textbf{34.18} & \textbf{38.65} & 62.6 & 66.7 \\
      \bottomrule
   \end{tabularx}
   \label{table:main}
\end{table*}
We comprehensively compare our \textsc{UAHOI} with the recently leading approaches in two representative human-object interaction datasets, HICO-DET~\citep{chao2018learning} and V-COCO~\citep{gupta2015visual} in Table\ref{table:main}. Additionally, we provide some qualitative results in Figure~\ref{fig:3}.

\subsection{Results for HICO-DET}
\noindent\textbf{Dataset.} We first assess \textsc{UAHOI} on human-object interaction datasets HICO-DET~\citep{chao2018learning}. HICO-DET has $47,776$ images, with $38,118$ for training and $9,658$ designated for testing. There are $600$ HOI categories (in total) over $117$ interactions and $80$ object categories. The interactions are further split into $138$ Rare and $462$ Non-Rare categories. We calculate the mAP scores using two different setups: (i) the Default Setup, where we compute the mAP across all test images; and (ii) the Known Object Setup, where we calculate the Average Precision (AP) for each object separately, only within the subset of images that contain the specified object.

\noindent\textbf{Evaluation$_{\!}$ Metric.} Following the standard evaluation protocoles ~\citep{qi2018learning,kim2021hotr}, we adopt mean Average Precision (mAP) as the primary metric for accessing model performance. 

\noindent\textbf{Comparison with State-of-the-Art Methods}
We first conduct experiments on HICO-DET~\citep{chao2018learning} with ResNet-50 as the backbone to verify the effectiveness of the proposed method, and report results in Table~\ref{table:main}. Compared to most transformer-based single-decoder works HOITrans~\citep{zou2021end} and QPIC~\citep{tamura2021qpic}, Our \textsc{UAHOI} achieves better performance, which validates the effectiveness of adopting multi-decoders for detecting more accurate Human-Object Interaction pairs. Comapred to RLIP~\citep{yuanrlip} which introduce language as additional cues for more accurate HOI detection, our method attains improvements from 32.84 mAP to 34.19 mAP for full evaluation under default setting. Even when comparing to the state-of-the-art method GEN-VLKT~\citep{liao2022gen}, our \textsc{UAHOI} reaches 34.19/31.54/35.27 mAP on the full/rare/non-rare evaluation under the default setting (The best results are highlighted in bold). Especially, \textsc{UAHOI} significantly promotes mAP from 29.25 to 31.54 for rare evaluation under default setting. These results substantiate our motivation to refine the human/object localization and interaction recognition via uncertainty estimation. For the known objects setting, it could be seen that \textsc{UAHOI} achieves 37.44/34.18/38.65 mAP on the full/rare/non-rare evaluation.

\subsection{Results for V-COCO}
\noindent\textbf{Dataset.$_{\!}$$_{\!}$} We next assess \textsc{UAHOI} on a smaller dataset V-COCO~\citep{gupta2015visual} which is originates from COCO~\citep{lin2014microsoft}. V-COCO has $2,533/2,867/4,946$ images for training, validation, and testing respectively. It consists of $80$ objects identical to those in HICO-DET and 29 action categories in total.

\noindent\textbf{Evaluation$_{\!}$ Metric.$_{\!}$} We also adopt Average Precision (AP) to report performance and compute it under two scenarios to address the challenge of objects missing due to occlusion. We denote these two scenarios with the subscripts $S_{role}^{S1}$ and $S_{role}^{S2}$. In scenario $S_{role}^{S1}$, when an object is occluded, we predict empty object boxes to consider the detected pair as a match with the corresponding ground truth. In scenario $S_{role}^{S2}$, object boxes are automatically considered matched in cases of occlusion, without the need to predict empty boxes.

\noindent\textbf{Comparison with State-of-the-Art Methods}
We next access \textsc{UAHOI} on V-COCO~\citep{gupta2015visual} dataset. Table~\ref{table:main} illustrates the results with both Scenario1 and Scenario2. It could be seen that \textsc{UAHOI} outperforms all existing methods without extra knowledge. We outperform state-of-the-art method GEN-VLKT~\citep{liao2022gen} by a large marin of 3.2 mAP under Scenario2. In addition, Compared to the methods with extra language knowledge, \textsc{UAHOI} is still competitive.

\subsection{Qualitative Results}
\begin{figure*}[t]
  \vspace{-5pt}
  \begin{center}
      \includegraphics[width=0.98\linewidth, height=200pt]{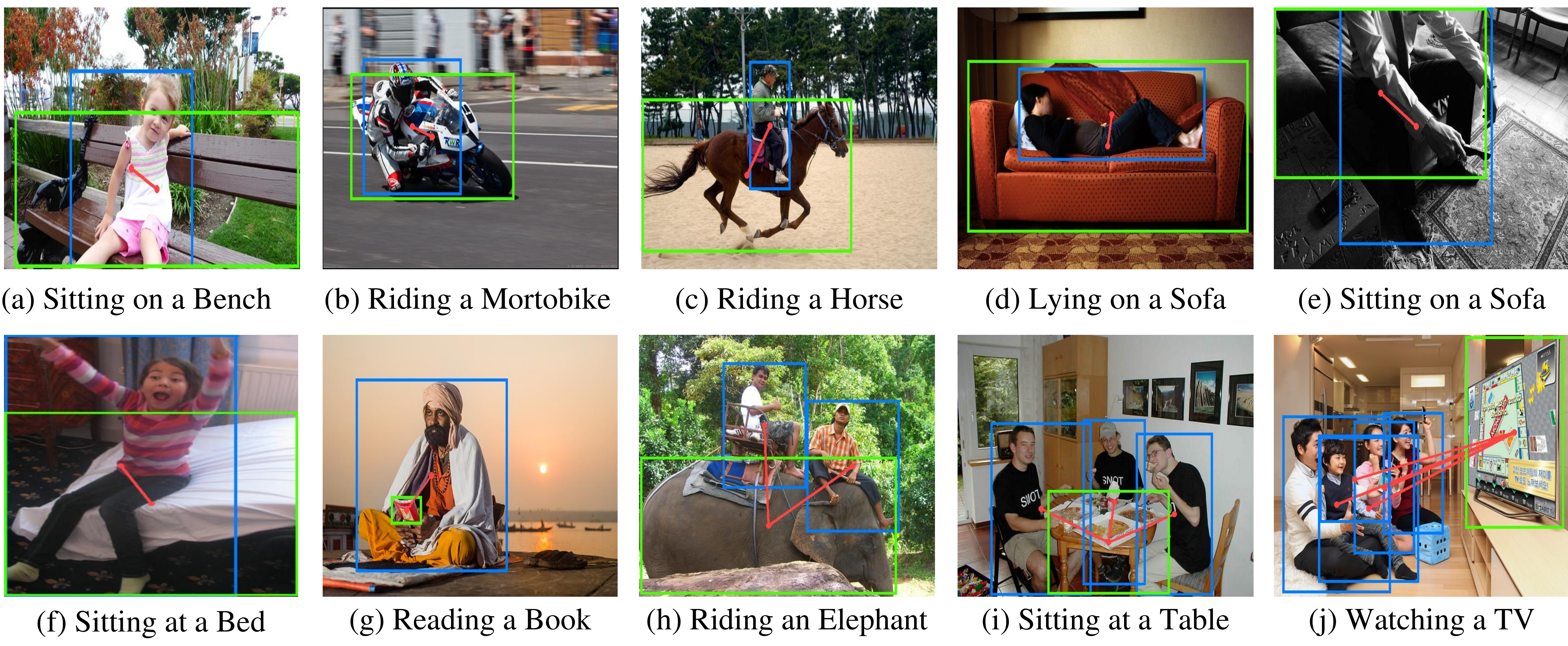}
      \end{center}
  \vspace{-15pt}
  \caption{Visualization results of our \textsc{UAHOI}.}
  \label{fig:3}
  \vspace{-5pt}
\end{figure*}

Additional qualitative results are presented in this section. As shown in Fig~\ref{fig:3}, we use red lines to denote the detected HOI pairs and blue/green boxes to represent human/object. It can be observed that for common interactions such as sitting, riding, lying, reading, etc., \textsc{UAHOI} demonstrates robust performance. Furthermore, when tackling more challenging scenes with an increased number of humans, ranging from 2 to 5 such as (h), (i), (j), \textsc{UAHOI} continues to perform effectively. 

\subsection{Ablation Study}
We evaluate the contribution of each component present in our framework. Specifically, we evaluate \textsc{UAHOI} on the task of HICO-DET~\citet{chao2018learning}, with Res-Net 50 backbone.

\begin{figure}[t]
  \vspace{-5pt}
  \begin{center}
      \includegraphics[width=0.98\linewidth, height=230pt]{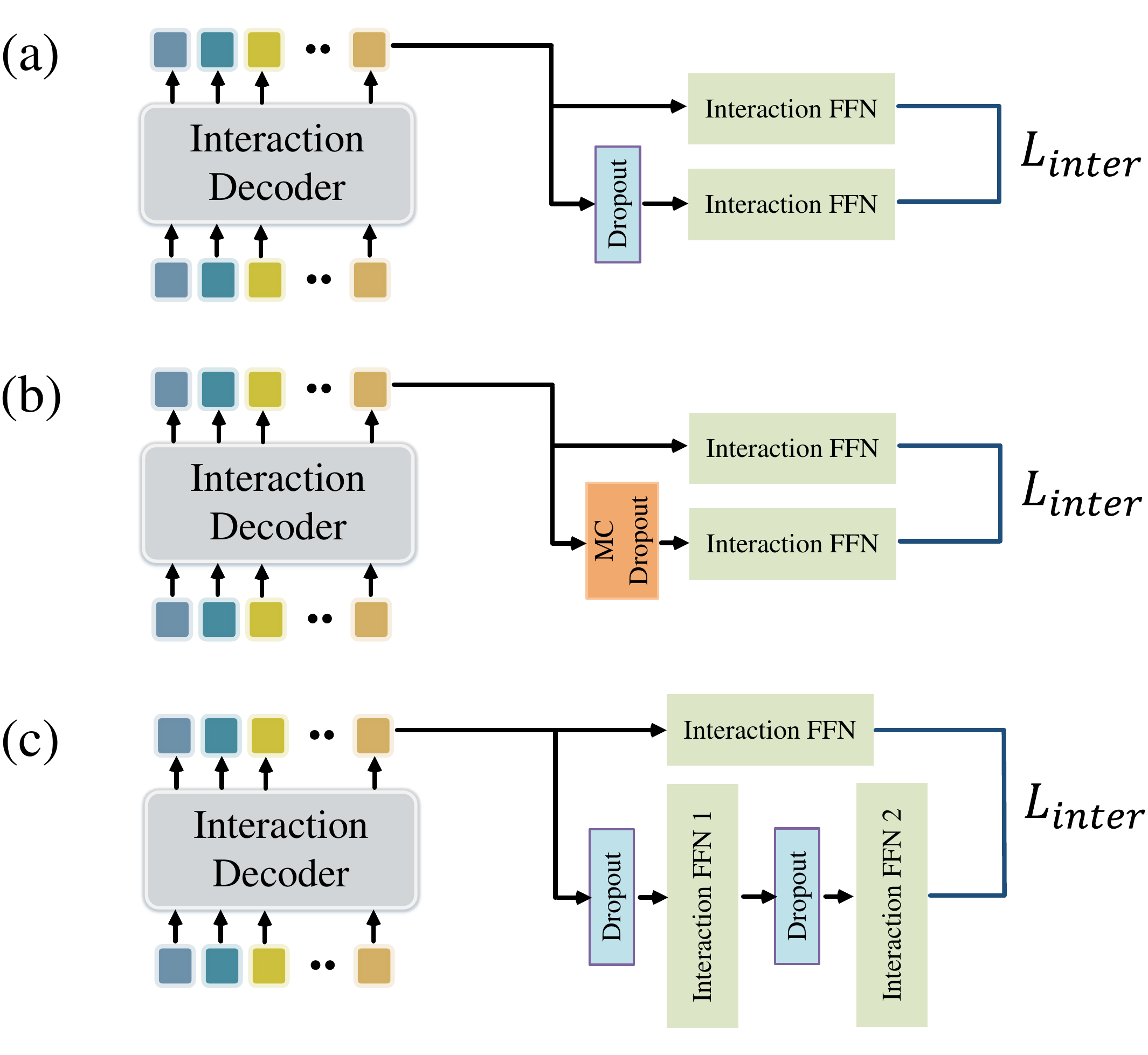}
      \end{center}
  \vspace{-15pt}
  \caption{For a more comprehensive validation of the effects of different uncertainty estimation methods, we further designed our interaction prediction tests by incorporating MC dropout (b), as well as utilizing architectures of Fully Feed-Forward Networks (FFN) with varying depths (c). By adding dropout at different layers, we achieved varying degrees of prediction variance. The results, comparisons, and further analyses are presented in Table~\ref{tab:strategy}.}
  \label{fig:4}
  \vspace{-4pt}
\end{figure}

\noindent\textbf{Major Components Analysis.} In this section, we conduct experiments for evaluating major components in our \textsc{UAHOI}: Uncertainty-aware Instance Localization and Interaction Refinement. As shown in Table~\ref{tab:kv}, our base model, utilizing a traditional handcrafted threshold, achieved 31.75 mAP. We then modeled the uncertainty of the bounding box and incorporated the proposed localization refinement module, which improved our results from 31.75 mAP to 32.52 mAP. Furthermore, to enhance the model accuracy in predicting complex interactions and prevent the model from discarding uncertain interaction categories due to a fixed threshold, we applied specific regularization to the prediction variance of the interactions. Using such a dynamic threshold, we optimized interaction uncertainty and achieved higher performance, reaching 33.65 mAP. Finally, by simultaneously employing both localization and interaction uncertainty modules, we elevated the results from 31.75 mAP to 33.65 mAP, achieving optimal performance. This validates the effectiveness of our approach in modeling uncertainty on two levels and integrating uncertainty regularization into our optimization objectives.

\begin{table}[t]\small
  \setlength{\belowcaptionskip}{2pt}
  \caption{Major component analysis on the HICO-DET test set. Results are averaged across three runs.}
  \label{tab:kv}
  \begin{tabularx}{\linewidth}{c|X|X|X}
    \bottomrule
    \hline
    Strategy & Full & Rare & Non-Rare \\
    \hline
    fixed threshold & 31.75 & 29.42 & 32.50 \\
    + localization refine & 32.52 & 30.48 & 33.63 \\
    + interaction refine & 33.65 & 31.27 & 34.88 \\
    \hline
    + both & $\mathbf{34.19}$ & $\mathbf{31.54}$ & $\mathbf{35.27}$ \\
    \shline
  \end{tabularx}
\end{table}

\noindent\textbf{Effect of various uncertainty.} Neural networks are theoretically capable of providing estimates of both confidence, known as aleatoric uncertainty, and model-based uncertainty, referred to as epistemic uncertainty. Existing literatures~\citep{gal2016dropout, zheng2021rectifying} employs various strategies to model uncertainty, and we compared our method with two other approaches. The first approach involves an extra classifier. This method adds another classifier to the existing network architecture to predict interactions, commonly referred to as the auxiliary classification head, while the original classifier is termed the primary classification head. Both classifiers provide predictions, and due to the inherent uncertainty of the model, the outputs from both classifiers are often more uncertain in challenging scenarios. Through regularization of both classifiers' predictions, refinement of interaction predictions is achieved. As shown in the table, this method reached an accuracy of 33.86 mAP. Subsequently, we compared this with Monte Carlo Dropout (MC-Dropout). MC-Dropout~\citep{gal2016dropout,lakshminarayanan2017simple,ciosek2019conservative} is used as another means to estimate epistemic uncertainty.  We implement different MC-Dropout rates of 0.5, 0.7, and 0.9 instead of the standard dropout. The results indicate that MC-Dropout also enhances performance and is not sensitive to the dropout rate.

\begin{table}[t]\small
  \setlength{\belowcaptionskip}{2pt}
  \caption{The mAP of different uncertainty modeling strategy on the HICO-DET test set. }
  \label{tab:strategy}
  \begin{tabularx}{\linewidth}{c|X|X|X}
    \bottomrule
    \hline
    Strategy & Full & Rare & Non-Rare \\
    \hline
    base & 32.52 & 30.48 & 33.63 \\
    + Additional Classifier & 33.86 & 30.79 & 34.82 \\
    + MC Dropout 0.5 & 33.15 & 33.65 & 34.79 \\
    + MC Dropout 0.7 & 33.22 & 33.71 & 34.70 \\
    + MC Dropout 0.9 & 32.97 & 33.51 & 34.35 \\
    + Dropout & $\mathbf{34.19}$ & $\mathbf{31.54}$ & $\mathbf{35.27}$ \\
    \shline
  \end{tabularx}
\end{table}

\begin{table}[t]\small
  \setlength{\belowcaptionskip}{2pt}
  \caption{The mAP of different dropout depth on the HICO-DET test set. }
  \label{tab:dropout}
  \begin{tabularx}{\linewidth}{c|X|X|X}
    \bottomrule
    \hline
    Strategy & Full & Rare & Non-Rare \\
    \hline
    base & 32.52 & 30.48 & 33.63 \\
    with one FFN layer & 34.19 & 31.54 & 35.27 \\
    \hline
    with two FFN layers & $\mathbf{34.27}$ & $\mathbf{31.60}$ & $\mathbf{35.43}$ \\
    \shline
  \end{tabularx}
\end{table}

\noindent\textbf{Effect of dropout depth.} 
According to the experiments presented in the Table~\ref{tab:strategy}, our model is not sensitive to changes in the dropout rate. Therefore, in this section, we perform dropout sampling on models with varying depths of Feed-Forward Networks (FFNs), with results shown in the Table~\ref{tab:dropout}. Our base model uses a single layer of FFN without employing dropout, achieving a result of 32.52 mAP. After implementing dropout and conducting uncertainty refinement, the score increased to 34.19 mAP. Using two layers of FFN and applying dropout to each layer further enhanced the performance to 34.27 mAP.

\noindent\textbf{Parameter sensitivity analysis on loss weights.} We conduct sensitivity analysis on the parameters of loss weights to evaluate the sensitivity of \textsc{UAHOI} on HICO-DET test set. As shown in Table~\ref{tab:lambda1}, we select loss weights $\lambda_{\text{o}}$ and $\lambda_{\text{a}}$  $ \in\{0.01,0.1,0.5,1.0,2.0\}$. When $\lambda_{\text{o}}$ and $\lambda_{\text{a}}$ change significantly, the model exhibits a bit sensitive to the assigned weights. However, when the changes in the coefficients are relatively small, the model is insensitive to the weights, and our method has achieved competitive results under various weights.

\begin{table}[t]\small
  \setlength{\belowcaptionskip}{2pt}
  \caption{Parameter sensitivity analysis on the weight of localization uncertainty loss on HICO-DET test set. }
  \label{tab:lambda1}
  \begin{tabularx}{\linewidth}{c|X|X|X|c|X|X|X}
    \bottomrule
    \hline
    $\lambda_o$ & Full & Rare & Non-Rare & $\lambda_a$ & Full & Rare & Non-Rare\\
    \hline
     0.01 & 32.18 & 29.45 & 32.89 & 0.01 & 33.65 & 30.45 & 34.89\\
     0.1 & 33.29 & 30.60 & 34.06 & 0.1 & 33.77 & 30.86 & 34.11\\
     0.5 & 34.09 & 31.49 & 35.19 & 0.5 & 33.86 & 31.04 & 34.55\\
     \hline
     1.0 & $\mathbf{34.19}$ & $\mathbf{31.54}$ & $\mathbf{35.27}$ & 1 & $\mathbf{34.19}$ & $\mathbf{31.54}$ & $\mathbf{35.27}$ \\
     \hline
     2.0 & 32.25 & 30.22 & 34.01 & 2.0 & 31.95 & 29.77 & 33.68\\
    \shline
  \end{tabularx}
\end{table}

\begin{figure}[t]
  \vspace{+10pt}
  \begin{center}
      \includegraphics[width=0.85\linewidth, height=100pt]{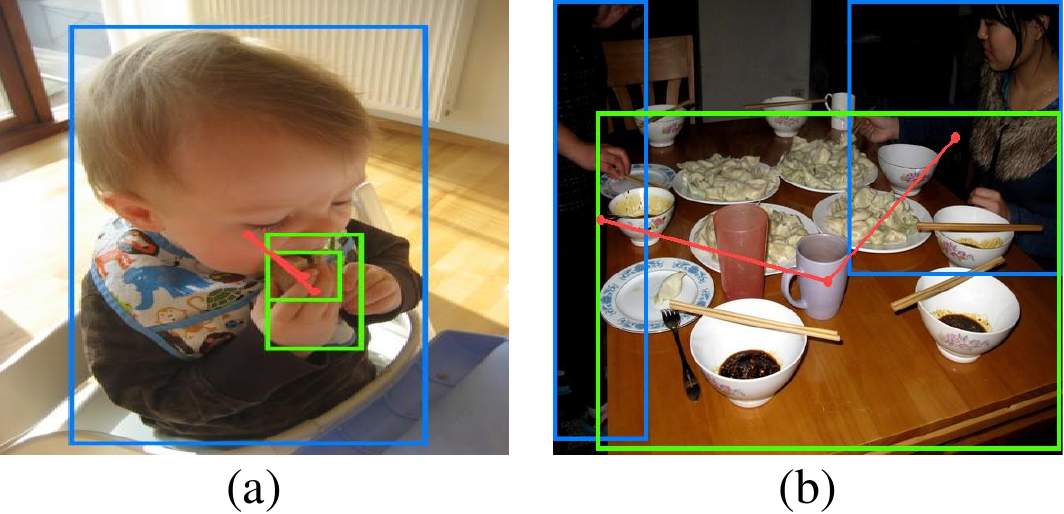}
      \end{center}
  \vspace{-10pt}
  \caption{Visualization results of two failure cases.}
  \label{fig:4}
  \vspace{-5pt}
\end{figure}

\section{Limitations and Future Works}Our architecture consists of a shared transformer encoder along with two separate decoders, making it computationally expensive, which could be detrimental in practical applications. To mitigate this impact, we employ a pre-trained backbone and only fine-tune our network during training to minimize the consumption of computational resources. Additionally, we show failure cases in Fig~\ref{fig:4}. Due to the presence of complex or overlapping objects, the model is unable to accurately identify all from the visual context.

In the future, the impact of dropout uncertainty at different positions (such as in deeper or shallower layers) in the network will be further explored and corresponding improvements will be made. Additionally, the method would be validated on the Scene Graph Generation (SGG) task, which could demonstrate the versatility and effectiveness of uncertainty estimation in other high-level semantic understanding tasks.

\section{Conclusions}
In this paper, we have delved into the application of uncertainty estimation within Human-Object Interaction (HOI) detection, exploring its integration in two key aspects: interaction and detection. For interaction, we utilized dropout not only as a regularization but also as a means for estimating the variance in interaction predictions. This dual-purpose use of dropout allows our model to assess and adapt to the reliability of its interaction classifications dynamically. For detection, we treated the bounding box coordinates as Gaussian-distributed random variables, which enables our system to quantify the uncertainty of object localizations and integrate this information into the learning process, thus enhancing prediction confidence and accuracy. Our approach is designed to be orthogonal to existing methods, allowing it to be seamlessly integrated with other techniques, thereby augmenting their effectiveness with robust uncertainty modeling capabilities. This integration capability provides a flexible framework that can be adopted to enhance current HOI detection systems without requiring extensive modifications to existing architectures.

\bibliographystyle{model2-names}
\bibliography{refs}

\end{document}